\documentclass[preprints,article,accept,moreauthors,pdftex]{Definitions/mdpi}

\firstpage{1} 
\pubvolume{xx}
\issuenum{1}
\articlenumber{5}
\pubyear{2019}
\copyrightyear{2019}
\history{Received: date; Accepted: date; Published: date}

\Title{LSTMs and Deep Residual Networks for Carbohydrate and Bolus Recommendations in\\
Type 1 Diabetes Management}

\Author{Jeremy Beauchamp $^{1}$\orcidA{}, Razvan Bunescu $^{2}$*, Cindy Marling $^{1}$, Zhongen Li $^{1}$, and Chang Liu $^{1}$}

\AuthorNames{Jeremy Beauchamp, Razvan Bunescu, Cindy Marling, Zhongen Li, and Chang Liu}

\address{%
$^{1}$ \quad School of Electrical Engineering and Computer Science, Ohio University; \{jb199113, marling, zl967813, liuc\}@ohio.edu\\
$^{2}$ \quad Department of Computer Science, University of North Carolina at Charlotte; rbunescu@uncc.edu}

\corres{Correspondence: rbunescu@uncc.edu; Tel.: +1-704-687-8444}

\abstract{To avoid serious diabetic complications, people with type 1 diabetes must keep their blood glucose levels (BGLs) as close to normal as possible. Insulin dosages and carbohydrate consumption are important considerations in managing BGLs. Since the 1960s, models have been developed to forecast blood glucose levels based on the history of BGLs, insulin dosages, carbohydrate intake, and other physiological and lifestyle factors. Such predictions can be used to alert people of impending unsafe BGLs or to control insulin flow in an artificial pancreas. In past work, we have introduced an LSTM-based approach to blood glucose level prediction aimed at "what if" scenarios, in which people could enter foods they might eat or insulin amounts they might take and then see the effect on future BGLs. In this work, we invert the "what-if" scenario and introduce a similar architecture based on chaining two LSTMs that can be trained to make either insulin or carbohydrate recommendations aimed at reaching a desired BG level in the future. Leveraging a recent state-of-the-art model for time series forecasting, we then derive a novel architecture for the same recommendation task, in which the two LSTM chain is used as a repeating block inside a deep residual architecture. Experimental evaluations using real patient data from the OhioT1DM dataset show that the new integrated architecture compares favorably with the previous LSTM-based approach, substantially outperforming the baselines. The promising results suggest that this novel approach could potentially be of practical use to people with type 1 diabetes for self-management of BGLs.}


\begin{document}

\section{Introduction and Motivation}


Diabetes self-management is a time-consuming, yet critical, task for people with type 1 diabetes.  To avoid serious diabetic complications, these individuals must continually manage their blood glucose levels (BGLs), keeping them as close to normal as possible.  They must avoid both low BGLs, or hypoglycemia, and high BGLs, or hyperglycemia, for their physical safety and well-being.  Diabetes self-management entails carefully monitoring BGLs throughout the day, by testing blood obtained from finger sticks and/or by using a continuous glucose monitoring (CGM) system.  It also entails making numerous daily decisions about the timing and dosage of insulin and the timing, ingredients, and quantity of food consumed.

Current diabetes self-management may be characterized as \emph{reactive}, rather than \emph{proactive}.  When BGLs are too high, individuals may take insulin to lower them, and when BGLs are too low, they may eat a snack or take glucose tablets to raise them.  The ability to accurately predict BGLs could enable people with type 1 diabetes to take preemptive actions \emph{before} experiencing the negative effects of hypoglycemia or hyperglycemia.  There have been efforts to model BGLs for the purpose of determining insulin dosages dating back to the 1960s \cite{boutayeb2016}.  There has been much recent work in BGL prediction for the purpose of providing support for diabetes self-management, including our own \cite{bunescu:icmla13,plis:maiha14}.  Accounts of some of the most recent BGL prediction efforts can be found in the proceedings of two international BGL prediction challenges \cite{kdh-2018-proceedings,kdh-2020-proceedings}.  It should be noted that, even with the benefit of accurate BGL predictions, individuals still need to determine how much to eat, how much insulin to take, and what other actions they can take to prevent hypoglycemia or hyperglycemia.

The broad goal of the research presented here is to essentially reverse the BGL prediction problem, and instead predict how many grams of carbohydrate (carbs) an individual should eat or how much insulin they should take in order to achieve a desired BGL target. We have previously introduced in  \cite{mirshekarian:embc19} an LSTM-based neural architecture that was trained to answer {\it what-if} questions of the type “What will my BGL be in 60 minutes if I eat a snack with 30 carbs 10 minutes from now?”. By using the BGL target as a feature and the carbohydrates or insulin as labels, we have shown in subsequent work \cite{beauchamp:kdh20} that a similar LSTM-based architecture can be trained instead to predict the number of carbs that should be consumed or the amount of insulin that should be taken during the prediction window in order to reach that BGL target. Preliminary results were reported in \cite{beauchamp:kdh20} only for the task of carbohydrate recommendation, where the aim was to achieve a desired target BGL 30 or 60 minutes into the future. The timing of the meal was variable within the prediction window and was used as one of the inputs to the model. In this paper, we update the task definition to make it more applicable to the type of situations that are frequently encountered in the self-management of type 1 diabetes. As such, the timing of the bolus or meal is now fixed at 10 minutes into the future, based on the assumption that patients are most interested in using the system right before making a meal or bolus decision. To achieve the desired BGL, the user can specify any time horizon between 30 and 90 minutes, giving them more flexibility in terms of how fast they want their BGL to change. Furthermore, we improve the LSTM-based architecture from \cite{beauchamp:kdh20} and use it as a repeating residual block in a deep residual forecasting network derived from the BGL prediction architecture recently proposed by \citet{rubin_falcone:nbeats_bgl}. The neural architecture from \cite{rubin_falcone:nbeats_bgl} is in turn a modified version of the N-BEATS architecture \cite{oreshkin:nbeats} that was shown to obtain state-of-the-art results on a wide array of time series forecasting problems. Overall, the new recommendation approach using the integrated deep network is generic in the sense that it can be trained to make recommendations about any variable that can impact BGLs, in particular, carbohydrates and insulin. Carbohydrate recommendations are potentially useful when someone wants to prevent hypoglycemia well in advance or when someone wants to achieve a higher target BGL before physical exercise that is expected to lower it. Bolus recommendations are useful prior to meals and also for lowering BGLs when individuals experience hyperglycemia.

The rest of the paper is organized as follows: Section~\ref{sec:related} presents related work in blood glucose level prediction and automated bolus calculators, and positions our work in relation to recent developments in these areas. Three major recommendation scenarios are introduced in Section~\ref{sec:scenarios}, followed in Section~\ref{sec:models} by a description of a set of baseline models and neural architectures that are designed to make recommendations in each of these scenarios. Section~\ref{sec:dataset} introduces the OhioT1DM dataset and explains how recommendation examples were derived from it. The experimental methodology and results are presented in Sections~\ref{sec:methodology} and~\ref{sec:results}, respectively. The paper concludes in Section~\ref{sec:conclusion} with a summary of our contributions and ideas for future work.

\section{Related Work}
\label{sec:related}

Bolus calculators have been in use since 2003 \cite{gross:calculator}, wherein a standard formula is used to calculate a recommended insulin dosage, taking into count parameters such as carbohydrate intake, carbohydrate-to-insulin ratio, insulin on board, and target BGL. \citet{walsh:jdst18} discuss some major sources of errors and the potential improvements that could be made to bolus advisors. One such improvement mentioned was the utilization of the massive amount of clinical data that is collected from these bolus advisor systems. AI techniques have been used to take advantage of this data to create more intelligent and personalized insulin recommendation systems. \citet{pesl:case_based} describe a bolus advisor system based on case-based reasoning that personalizes an individual's bolus calculations. The system gathers simple information, such as the range of recent blood glucose levels and the time of day, and compares the current situation to situations from the past to find similar cases. The system then uses the bolus recommendation from a similar previous case and adapts it to the current scenario. The work by \citet{tyler:knn} shows a K-nearest-neighbors based system that provides weekly recommendations to improve the effectiveness of an individual's multiple daily injection therapy. With the amount of clinical data collected from CGM systems and wearable sensors, deep learning is a natural fit for insulin advisor systems. The work in \citet{stavroula:dtt} represents an early attempt at creating insulin recommendations by utilizing neural networks. \citet{cappon:jdst18} observe that the standard formula approach to bolus calculation ignores potentially important preprandial conditions, such as the glucose rate of change. To address this, they propose a simple feed-forward neural network that utilizes CGM data and other easily accessible factors to personalize the bolus calculation. The experimental evaluations on simulated data show a small, but statistically significant improvement in the blood glucose risk index. Simulated data is also used by \citet{sun:jbhi19}, where a basal-bolus advisor is trained using reinforcement learning in order to provide personalized suggestions to people with type 1 diabetes taking multiple daily injections of insulin. \citet{zhu:bolus_drl} also utilize simulated type 1 diabetes data for their deep reinforcement learning approach to personalizing bolus calculations. They use the UVA/Padova type 1 diabetes simulator \cite{dalla_man:simulator} to train models to learn how to reduce or amplify the bolus dose recommended by a bolus calculator to provide more personalized recommendations.

In contrast with previous work that used a type 1 diabetes simulator \cite{cappon:jdst18,sun:jbhi19,zhu:bolus_drl}, the systems described in this paper are trained and evaluated on data acquired from people with type 1 diabetes, derived from the OhioT1DM dataset \cite{ohiot1dm:marling:kdh18} as explained in Section~\ref{sec:dataset}. The case-based reasoning system introduced by \citet{pesl:case_based} also makes use of real-patient data; however, their system does not learn directly from this data. Instead, it learns from clinical experts' advice, requiring also that the system is tweaked on a regular basis. Elsewhere \cite{mirshekarian:embc19}, we have shown that patterns that are learned from simulated data do not always transfer to real data, and vice-versa. By training and evaluating on real data, the results reported in this paper are expected to be more representative of the actual performance of the recommendation system if it were to be deployed in practice. Most of the related work on bolus recommendations presented above use global means of evaluating system performance, such as the percentage of time BGLs are in a target range \cite{sun:jbhi19,zhu:bolus_drl}, or the blood glucose risk index \cite{cappon:jdst18}. In contrast, our approaches are evaluated retrospectively on how close their recommendations are to the actual carbohydrate content or bolus dosages that led to a particular target BGL. As such, the trained models can be used to make recommendations in order to achieve a specific BGL. The neural architectures that we propose in this paper are also general in the sense that they can be used to make recommendations for any type of discrete intervention that may impact BGLs. While, in this paper, they are evaluated on bolus, carbs, and bolus given carbs recommendations, we also see them as applicable for recommending other relevant variables, such as exercise.


\section{Three Recommendation Scenarios}
\label{sec:scenarios}

We assume that blood glucose levels are measured at 5 minute intervals through a CGM system. We also assume that discrete deliveries of insulin (boluses) and continuous infusions of insulin (basal rates) are recorded. Subjects provide the timing of meals and estimates of the number of grams of carbohydrate in each meal. Given the available data up to and including the present (time $t$), the system aims to estimate how much a person should eat or bolus 10 minutes from now (time $t+10$) such that their blood glucose will reach a target level $\tau$ minutes after that action (time $t + 10 + \tau$). A system that computes these estimates could then be used in the following three recommendation scenarios:
\begin{enumerate}
    \item {\bf Carbohydrate Recommendations}: Estimate the amount of carbohydrate $C_{t+10}$ to have in a meal in order to achieve a target BG value $G_{t+10+\tau}$.
    \item {\bf Bolus Recommendations}: Estimate the amount of insulin $B_{t+10}$ to deliver with a bolus in order to achieve a target BG value $G_{t+10+\tau}$.
    \item {\bf Bolus Recommendations given Carbohydrates}: Expecting that a meal with $C_{t+20}$ grams of carbohydrate will be consumed 20 minutes from now, estimate the amount of insulin $B_{t+10}$ to deliver with a bolus 10 minutes before the meal in order to achieve a target BG value $G_{t+10+\tau}$.
\end{enumerate}
These recommendation scenarios were designed to align with decision-making situations commonly encountered by people with type 1 diabetes. In particular, the corresponding recommendation systems would help an individual to estimate how much to eat or bolus for the purpose of raising or lowering their BGL (scenarios 1 and 2), as well as how much to bolus for a planned meal (scenario 3). 


In the following Section~\ref{sec:models}, we describe a number of baseline models and neural architectures, all implementing the three types of recommendations. The neural architectures use Long Short-Term Memory (LSTM) networks either in a standalone prediction model (Section~\ref{sec:lstm}) or integrated as basic repeating blocks in a deep residual network (Section~\ref{sec:nbeats}). The models are trained on examples extracted from the OhioT1DM dataset~\cite{ohiot1dm:marling:kdh18}, as explained in Section~\ref{sec:dataset}. Ideally, to match the intended use of these recommendations in practice, training examples should not have any extra meals or boluses in the prediction window $[t, t + 10 + \tau]$. Following the terminology from \cite{mirshekarian:embc19}, we call these examples {\it inertial}. However, to benefit from a larger number of training examples, we also train and evaluate models on a more general class of {\it unrestricted} examples, in which other bolus or meal events are allowed to appear in the prediction window. Correspondingly, experimental results for inertial vs. unrestricted examples are presented in Section~\ref{sec:results}.

\section{Baseline Models and Neural Architectures}
\label{sec:models}

Given training data containing time series of blood glucose levels, meals with their carbohydrate intake, and boluses with their corresponding insulin dosages, we define the following two baselines:
\begin{enumerate}
    \item {\bf Global average}: 
    For the carbohydrate recommendation scenario, the average number $\mu$ of carbs over all of the meals in the subject's training data is computed and used as the estimate for all future predictions for that subject, irrespective of the context of the example. Analogously, for the bolus and bolus given carbs recommendation scenarios, $\mu$ is the average amount of insulin dosage over all boluses in the subject's training data. This is a fairly simple baseline, as it predicts the same average value for every test example for a particular subject.
    
    \item {\bf ToD average}: In this Time-of-Day (ToD) dependent baseline, an average number of carbs or an average amount of bolus insulin is computed for each of the following five time windows during a day:
	\begin{itemize}
		\item 12am-6am: $\mu_1$ = early breakfast / late snacks.
		\item 6am-10am: $\mu_2$ = breakfast.
		\item 10am-2pm: $\mu_3$ = lunch.
		\item 2pm-6pm: $\mu_4$ = dinner.
		\item 6pm-12am: $\mu_5$ = late dinner / post-dinner snacks.
	\end{itemize}
    The average for each ToD interval is calculated over all of the meals or boluses appearing in the corresponding time frame in the subject's training data. At test time, to make a recommendation for time $t+10$, we first determine the ToD interval that contains $t+10$ and output the corresponding ToD average.
\end{enumerate}
Given sufficient historical data, the ToD baseline is expected to perform well for individuals who tend to eat very consistently and have regular diets. However, it is expected to perform poorly for individuals who have a lot of variation in their diets.


\begin{figure*}[t]
    \includegraphics[width=\textwidth]{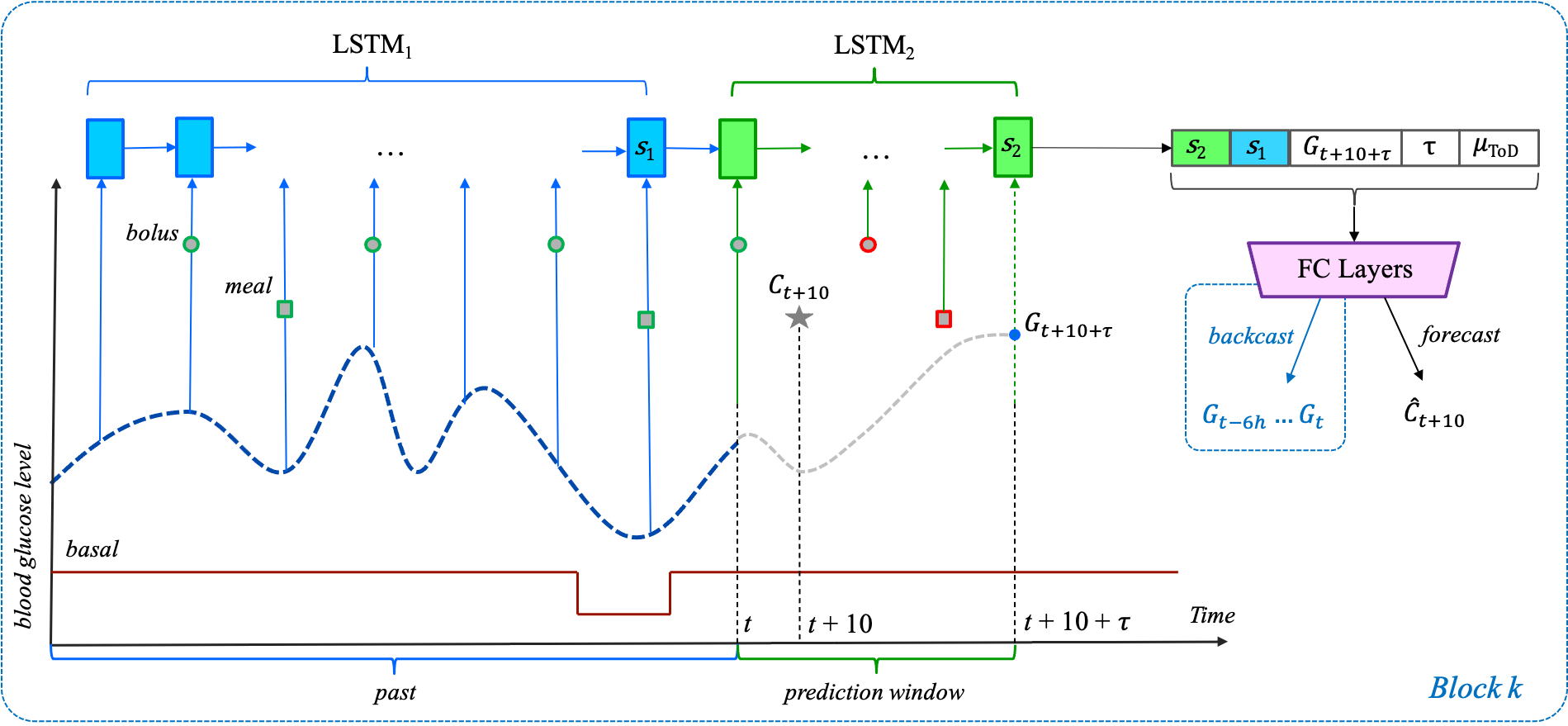}
    \caption{The general neural network architecture for the carbohydrate recommendation scenario. The dashed blue line in the graph represents a subject's BGL, while the solid brown line represents the basal rate of insulin. The gray star represents the meal at time $t+10$. The other meals are represented by squares, and boluses are represented by circles. Meals and boluses with a red outline cannot appear in {\it inertial} examples, but are allowed in {\it unrestricted} examples. The blue units in $\text{LSTM}_{1}$ receive input from different time steps in the past. The green units in $\text{LSTM}_{2}$ receive input from the prediction window. The purple trapezoid represents the 5 fully connected layers, whereas the output node at the end computes the prediction.}
    \label{fig:carbs}
\end{figure*}

\begin{figure*}[t]
    \includegraphics[width=\textwidth]{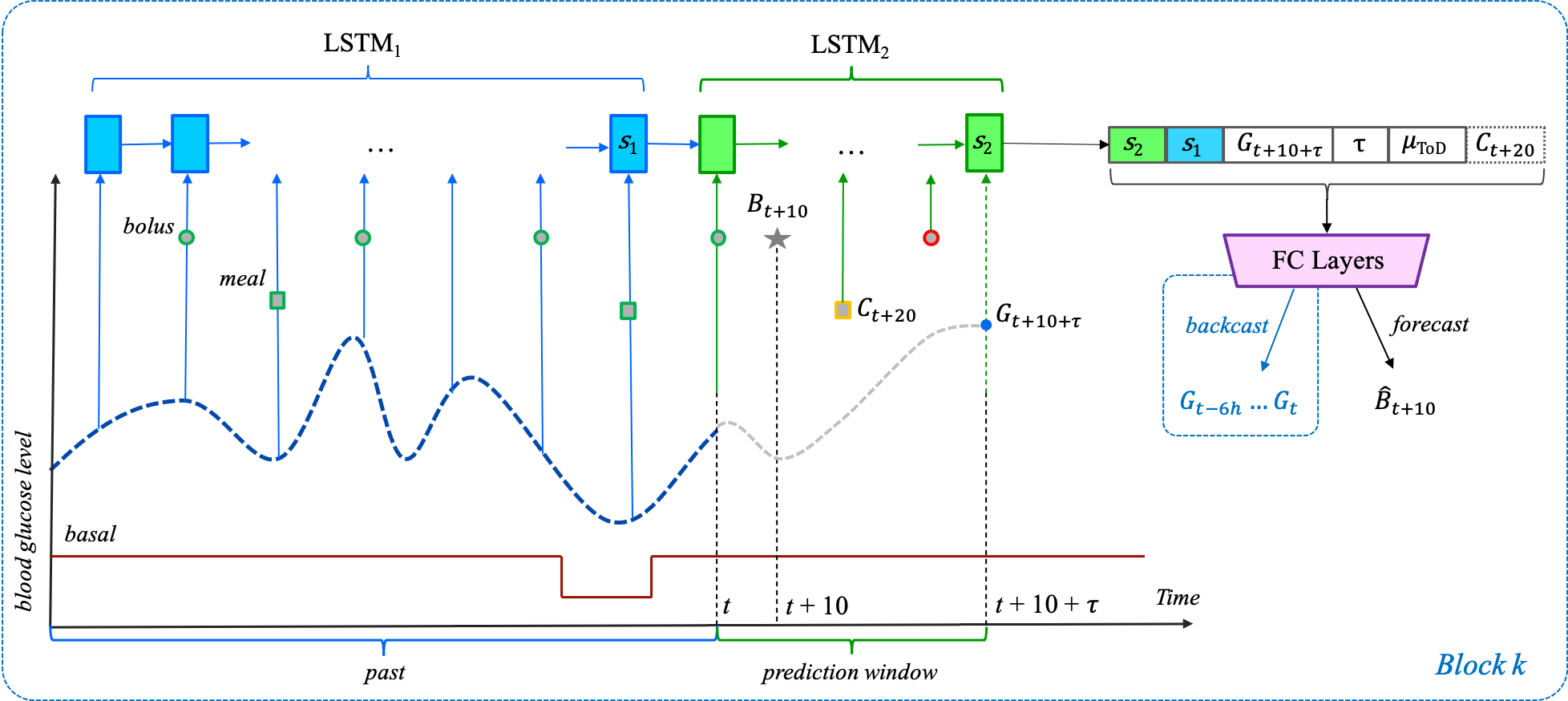}
    \caption{The general neural network architecture for the bolus and bolus given carbs recommendation scenarios. The architecture itself is similar to that shown in Figure~\ref{fig:carbs}. The grey star now represents the bolus at time $t+10$. For the bolus recommendation scenario, the events outlined in red or orange are not allowed in {\it inertial} examples.  However, in the bolus given carbs scenario, the meal event $C_{t+20}$ shown with the yellow outline is an important part of each example, be it inertial or unrestricted. As such, in this scenario, the dashed $C_{t+20}$ becomes part of the input to the FCN.}
    \label{fig:bolus}
\end{figure*}

\begin{figure*}[t]
    \includegraphics[width=\textwidth]{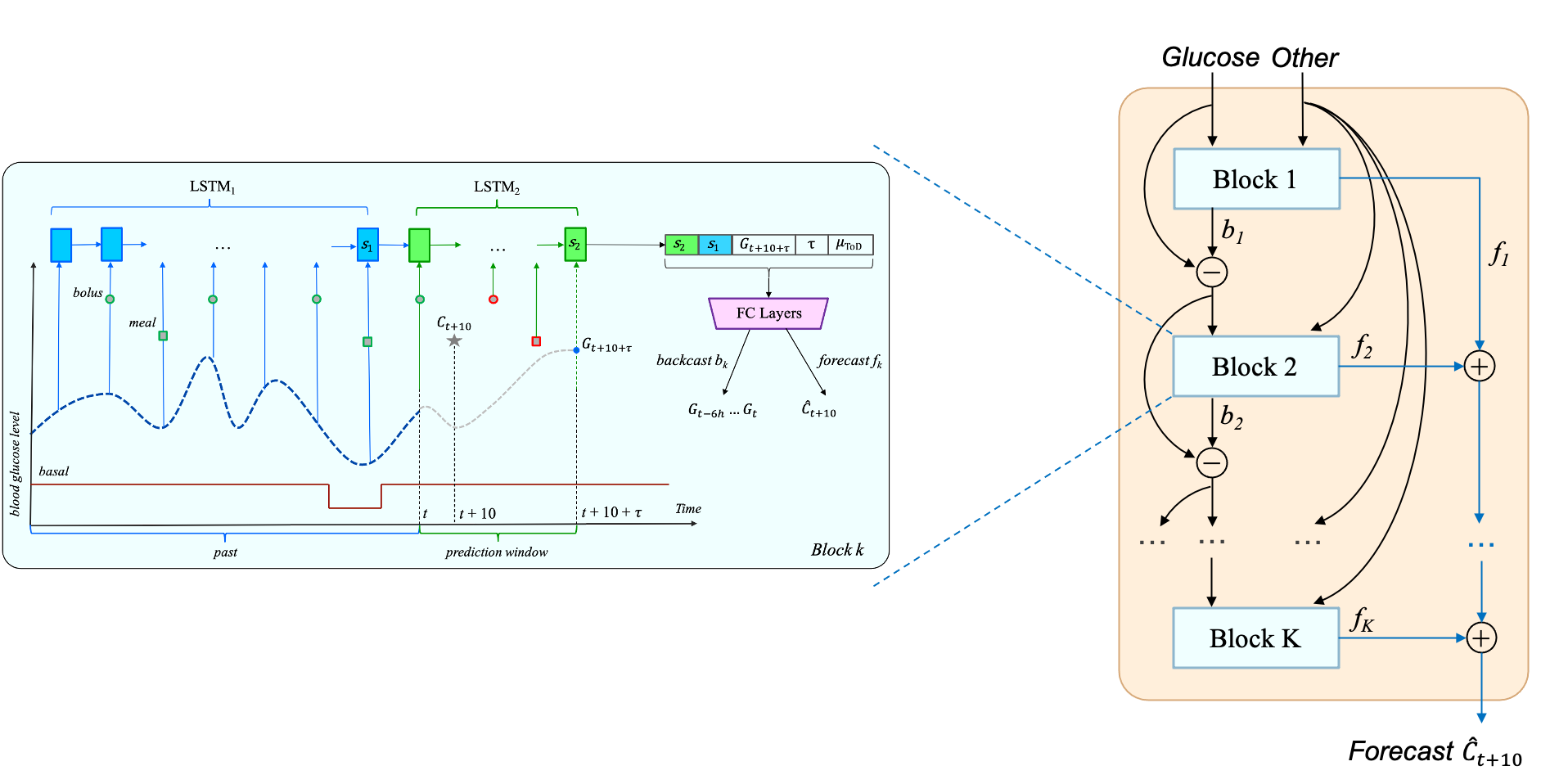}
    \caption{The N-BEATS inspired deep residual architecture for carbohydrate recommendation. A similar architecture is used for bolus and bolus given carbs recommendations.}
    \label{fig:nbeats}
\end{figure*}

\subsection{LSTM Models for Carbohydrate and Insulin Recommendation}
\label{sec:lstm}

While simple to compute and use at test time, the two baselines are likely to give suboptimal performance, as their predictions ignore the history of BGL values, insulin (boluses and basal rates), and meals, all of which could significantly modulate the effect a future meal and/or bolus might have on the BGL. To utilize this information, we first introduce the general LSTM-based network architecture shown in Figure~\ref{fig:carbs} for carb recommendation and Figure~\ref{fig:bolus} for bolus recommendation. The first component in the architecture is a recurrent neural network instantiated using Long Short-Term Memory (LSTM) cells \cite{hochreiter:nc97}, which is run over the previous 6 hours of data, up to and including the present time $t$. At each time step (every 5 minutes), this LSTM$_1$ network takes as input the BGL, the carbs, and the insulin dosages recorded at that time step. While sufficient for processing {\it inertial} examples, the same LSTM cannot be used to process events that may appear in the prediction window $(t, t+10+\tau)$ of {\it unrestricted} examples, because BGL values are not available in the future. Therefore, when training on unrestricted examples, the final state computed by the LSTM$_1$ model at time $t$ is projected using a linear transformation and used as the initial state for a second LSTM model, LSTM$_2$, that is run over all the time steps in the prediction window $(t, t+10+\tau)$. The final state computed by LSTM$_1$ (for inertial examples) is appended to the final state computed by LSTM$_2$ (for unrestricted examples) and is then used as input to a fully connected network (FCN) whose output node computes an estimate of the carbs or bolus insulin at time $t+10$. In addition to the LSTM final state(s), the input to the FCN contains the following features:
\begin{itemize}
    \item The target blood glucose level $\tau+10$ minutes into the future, i.e., $G_{t + 10 + \tau}$.
    \item The prediction horizon $\tau$.
    \item The ToD average for the time frame that contains $t+10$.
    \item For the bolus given carbs scenario only, the planned amount $C_{t + 20}$ of carbohydrate becomes part of the input, too.
\end{itemize}
Each LSTM uses vectors of size 32 for the states and gates, whereas the FCN is built with up to 5 hidden layers, each consisting of 64 ReLU neurons, and one linear output node. Note that by using the final state of LSTM$_1$  to initialize LSTM$_2$, the latter's final state should theoretically be able to capture any useful information that is represented in the final state of LSTM$_1$, which may call into question the utility of concatenating the two final states. This architectural decision is nevertheless supported empirically through evaluations on the validation data, which show improvements in prediction performance when both states are used (Section~\ref{sec:development}).

\subsection{Deep Residual Models for Carbohydrate and Insulin Recommendation}
\label{sec:nbeats}

\citet{oreshkin:nbeats} have recently introduced a new architecture for time series forecasting, the Neural Basis Expansion for Interpretable Time-Series Forecasting (N-BEATS). The basic building {\it block} of N-BEATS is a fully connected structure that initially takes as input a fixed-size {\it lookback period} of past values of the target variable and outputs both {\it forecast} (estimates of future values) and {\it backcast} (estimates of past values) vectors. Blocks are organized into {\it stacks} such that the backcast of the current block is subtracted from its input and fed as input to the next block, whereas the forecast vectors from each block are summed up to provide the overall {\it stack forecast}. The stacks themselves are chained in a pipeline where the backcast output of one stack is used as input for the next stack. The overall model forecast is then computed by accumulating the forecasts across all the stacks.

The N-BEATS architecture was shown in \cite{oreshkin:nbeats} to obtain state-of-the-art performance on a wide range of time series prediction tasks, which suggests that it can serve as a model of choice for BGL prediction, too. However, in BGL prediction, time series of variables other then the primary blood glucose are also available. Correspondingly, \citet{rubin_falcone:nbeats_bgl} changed the N-BEATS block architecture to also use as input secondary, sparse variables such as meals and bolus insulin, while still backcasting only on the primary forecasting variable, blood glucose. To account for the temporal nature of the input, the fully connected structure of the basic N-BEATS block was replaced with LSTMs, followed by one fully connected layer whose output was split into the backcast and forecast vector. Additional per-block forecast and backcast loss terms were also added to provide more supervision.

We adapted the deep residual network from \cite{rubin_falcone:nbeats_bgl} to perform carb or bolus recommendations by using the LSTM-based architecture from Section~\ref{sec:lstm} to instantiate each block in the stack, as shown in Figure~\ref{fig:nbeats}. Compared to the architecture from \cite{rubin_falcone:nbeats_bgl}, the most significant differences are:
\begin{enumerate}
    \item The use of a chain of two LSTM networks in each block.
    \item The inclusion of additional inputs to the fully connected layers, i.e. the target BG level, the time horizon, and the ToD average.
    \item While backcasting is still done for blood glucose, forecasting is done for carbs or bolus, depending on the recommendation scenario.
\end{enumerate}
While \citet{oreshkin:nbeats} used 30 blocks and \citet{rubin_falcone:nbeats_bgl} used 10 blocks, the validation experiments for the recommendation tasks showed that the most effective deep residual architecture uses only up to 5 blocks, depending on the recommendation scenario (Section~\ref{sec:development}).

\section{Using the OhioT1DM Dataset for Recommendation Examples}
\label{sec:dataset}

To evaluate the proposed recommendation models, we create training and test examples based on data collected from 12 subjects with type 1 diabetes that is distributed with the OhioT1DM dataset~\cite{ohiot1dm:marling:kdh20}. The 12 subjects are partitioned in two subsets as follows:
\begin{enumerate}
    \item {\bf OhioT1DM 2018}: This is the first part of the dataset, containing data collected from 6 patients. It was used for the 2018 Blood Glucose Level Prediction (BGLP) challenge \cite{kdh-2018-proceedings}.
    \item {\bf OhioT1DM 2020}: This is the second part of the dataset, containing data collected from 6 additional patients. It was used for the 2020 BGLP challenge \cite{kdh-2020-proceedings}.
\end{enumerate}
Time series containing the basal rate of insulin, boluses, meals, and BGL readings were collected over 8 weeks, although the exact number of days varies from subject to subject. 
Insulin and BGL data was automatically recorded by each subject's insulin pump.  Meal data was collected in two different ways.  Subjects self reported meal times and estimated carbs via a smartphone interface.  Subjects also entered estimated carbs into a bolus calculator when bolusing for meals, and this data was recorded by the insulin pump.

\subsection{The Bolus Wizard}
\label{sec:bw}

To determine their insulin dosages, the subjects in the OhioT1DM study used a bolus calculator, or "Bolus Wizard (BW)," which was integrated in their insulin pumps.  They used it to calculate the bolus amount before each meal as well as when using a bolus to correct for hyperglycemia.  To use the BW, a subject enters their current blood glucose level and, if eating, their estimated number of grams of carbohydrate.  To calculate a recommended insulin dosage, the BW uses this input from the subject, plus the amount of active insulin the subject already has in their system, along with the following three pre-programmed, patient-specific parameters:
\begin{enumerate}
    \item The carb ratio, which indicates the number of grams of carbohydrate that are covered by a unit of insulin.
    \item The insulin sensitivity, which tells how much a unit of insulin is expected to lower the subject's blood glucose level.
    \item The target blood glucose range, which defines an individual's lower and upper boundaries for optimal blood glucose control.
\end{enumerate}
All three parameters may vary, for the same individual, throughout the day and over time\footnote{\url{https://www.medtronicdiabetes.com/loop-blog/4-features-of-the-bolus-wizard}}. Given this input and these parameters, the BW calculates the amount of insulin the subject should take to maintain or achieve a blood glucose level within their target range. The calculation is displayed to the subject as a recommendation, which the subject may then accept or override.

Based on the inputs and the patient-specific parameters described above, the BW uses a deterministic formula to calculate the bolus amount before each meal. As such, when trained in the bolus given carbs recommendation scenario, there is the risk that the deep learning models introduced in Section~\ref{sec:models} might simply learn to reproduce this deterministic dependency between bolus and carbs, while ignoring the target BG level that is used as input. However, this is not the case in our experimental settings, for the following reasons:
\begin{itemize}
    \item The ML models do not have access to any of the three patient-specific parameters above, which can change throughout the day and over time, and which are set based on advice from a health care professional.
    \item The BW uses a fixed target BG range depending on the time of day, whereas the target in the recommendation scenarios is a more specific BG level, to be achieved at a specific time in the near future.
    \item The amount of insulin calculated by the BW is only a recommendation, which is often overridden by subjects. We ran an analysis of the OhioT1DM dataset in which we counted how many times the amount of insulin that was actually delivered was different from the bolus recommendation. The analysis revealed that, of all the times that the BW was used, its recommendation was overridden for about a fifth of the boluses. Furthermore, there are subjects in the dataset who often did not use the BW (540 and 567), or who 
    chose to not use the BW at all (596).
\end{itemize}
Therefore, the ML models will have to go beyond using solely the carbohydrate amount in the intended meal. In order to fit the bolus recommendation examples, they will need to learn the impact that a bolus has on the target BG level for the specified prediction horizon, taking into account the amount of carbohydrate in the meal as well as the history of carbs, insulin, and BG levels. This data driven approach to bolus recommendation relieves the physician from the cognitively demanding task of regularly updating parameters such as the carb ratio and the insulin sensitivity, which often requires multiple fine tuning steps. In contrast, any relevant signal that is conveyed through the carb ratio and insulin sensitivity is expected to be learned by the ML models from the data.

\subsection{Pre-processing of Meals and BG Levels}
\label{sec:pre-processing}

While exploring the data, it was observed that self-reported meals and their associated boluses were in unexpected temporal positions relative to each other. For many meals, patients recorded a timestamp in the smartphone interface that preceded the corresponding bolus timestamp recorded in the insulin pump. This was contrary to what was recommended to the subjects by their physicians, which was to bolus shortly before the meal, and no more than 15 minutes prior to the meal. This discrepancy is likely due to subjects reporting incorrect meal times in the smartphone interface.

To correct the meal events, we used the data input to the BW in the insulin pump and ran a pre-processing step that changed the timestamp of each meal associated with a bolus to be exactly 10 minutes after that bolus. For these meals, we also used the number of carbs provided to the BW, which is likely to be more accurate than the estimate provided by the subject through the smartphone interface. To determine the self-reported meal event associated with a bolus having non-zero carb input, we searched for the meal that was closest in time to the bolus, within one hour before or after it. In case there were two meals that are equally close to the bolus, we selected the one for which the number of carbs from the smartphone interface was closest to the number of carbs entered into the BW. If no self-reported meal was found within one hour of the bolus, it was assumed that the subject forgot to log their meal on the smartphone interface. As such, a meal was added 10 minutes after the bolus, using the amount of carbs specified in the BW for that bolus. Ablation results reported in Section~\ref{sec:pre-evaluation} show that this pre-processing of meal events leads to significantly more accurate predictions, which further justifies the pre-processing.

All gaps in BGL data are filled in with linearly interpolated values. However, we filter out examples that meet any of the following criteria:
\begin{enumerate}
    \item The BGL target is interpolated.
    \item The BGL at present time $t$ is interpolated.
    \item There are more than 2 interpolated BGL measurements in the one hour of data prior to time $t$.
    \item There are more than 12 interpolated BGL measurements in the 6 hours of data prior to time $t$.
\end{enumerate}

\subsection{Mapping Prototypical Recommendation Scenarios to Datasets}
\label{sec:mapping}

According to the definition given in Section~\ref{sec:scenarios}, the carbohydrate recommendation scenario refers to estimating the amount of carbohydrate $C_{t+10}$ to have in a meal in order to achieve a target BG value $G_{t+10+\tau}$. This is done by using the history of data up to and including the present time $t$. However, many carbohydrate intake events $C_{t+10}$ are regular meals, which means that they are preceded by a bolus event at time $t$. Since in the carbohydrate recommendation scenario we are especially interested in scenarios where the subject eats in order to correct or prevent hypoglycemia, we created two separate datasets for carbohydrate prediction:
\begin{enumerate}
    \item Carbs$^{(\pm b)}$: this will contain examples for all carbohydrate intake events, with $(+b)$ or without $(-b)$ an associated bolus.
    \item Carbs$^{(-b)}$: this will contain examples only for carbohydrate intake events without $(-b)$ an associated bolus.
\end{enumerate}
Most of the Carbs$^{(-b)}$ examples are expected to happen in one of three scenarios: (1) when correcting for hypoglycemia; (2) before exercising; and (3) when having a bedtime snack to prevent nocturnal hypoglycemia. Given that they are only a small portion of the overall carbohdyrate events, in Section~\ref{sec:results} we present results for both Carbs$^{(\pm b)}$ and Carbs$^{(-b)}$ recommendation scenarios.

Furthermore, mirroring the two bolus recommendation scenarios introduced in Section~\ref{sec:scenarios}, we introduce the following notation for the corresponding datasets:
\begin{enumerate}
    \item Bolus$^{(\pm c)}$: this will contain examples for all bolus events, with $(+c)$ or without $(-c)$ an associated carbohydrate intake.
    \item Bolus$^{(+c)}$: this will contain examples only for the bolus events with $(+c)$ an associated carbohydrate intake.
\end{enumerate}
The three major recommendation scenarios introduced in Section~\ref{sec:scenarios} can then be mapped to the corresponding datasets as follows:
\begin{enumerate}
    \item {\bf Carbohydrate Recommendations}: Estimate the amount of carbohydrate $C_{t+10}$ to have in a meal in order to achieve a target BG value $G_{t+10+\tau}$.
    \begin{itemize}
        \item Carbs$^{(-b)}$, inertial: this reflects the prototypical scenario where a carbohydrate intake is recommended to correct or prevent hypoglycemia.
    \end{itemize}
    \item {\bf Bolus Recommendations}: Estimate the amount of insulin $B_{t+10}$ to deliver with a bolus in order to achieve a target BG value $G_{t+10+\tau}$.
    \begin{itemize}
        \item Bolus$^{(\pm c)}$, inertial: this reflects the prototypical scenario where a bolus is recommended to correct or prevent hyperglycemia. Because in the inertial case a carb event cannot appear after the bolus, this could also be denoted as Bolus$^{(-c)}$.
    \end{itemize}
    \item {\bf Bolus Recommendations given Carbohydrates}: Expecting that a meal with $C_{t+20}$ grams of carbohydrate will be consumed 20 minutes from now, estimate the amount of insulin $B_{t+10}$ to deliver with a bolus 10 minutes before the meal in order to achieve a target BG value $G_{t+10+\tau}$.
    \begin{itemize}
        \item Bolus$^{(+c)}$, inertial: this reflects the prototypical scenario where a bolus is recommended before a meal.
    \end{itemize}
\end{enumerate}

\subsection{Carbohydrate and Bolus Statistics}
\label{sec:statistics}

Table~\ref{tab:meals} shows the number of carbohydrate events in each subject's pre-processed data, together with the minimum, maximum, median, average, and standard deviation for the number of carbs per meal. Overall, the average number of carbs per meal is between 22 and 69, with the exception of subjects 570 and 544 whose meal averages and standard deviations are significantly larger. 
Table~\ref{tab:boluses} shows similar statistics for boluses and their dosages, expressed in units of insulin.
Overall, the number of boluses is more variable than the number of meals. There is also a fairly wide range of average bolus values in the data, with subject 567 having a much higher average than other subjects. It is also interesting to note that subject 570, who had the largest average carbs per meal, had more than twice the number of boluses than any other subject while at the same time having the lowest average bolus. Subject 570 also used many dual boluses, which we did not use as prediction labels because the scope of the project covers only recommendations for regular boluses.

\begin{table}[t]\setlength{\tabcolsep}{4pt}
\begin{center}
\caption{Per subject and total meal and carbohydrate per meal statistics: Minimum, Maximum, Median, Average, and Standard Deviation (StdDev). Carbs$^{(\pm b)}$ refers to all carbohydrate intake events; Carbs$^{(-b)}$ refers to carbohydrate intakes without a bolus. Statistics are shown for the 2018 subset, the 2020 subset, and for the entire OhioT1DM dataset.}
\label{tab:meals}
\begin{tabular}{|crr|rrrrc|}
    \cline{4-8}
    \multicolumn{3}{c}{} & \multicolumn{5}{|c|}{Carbs Per Meal}\\
	\hline
	Subject & \multicolumn{1}{c}{Carbs$^{(\pm b)}$} & \multicolumn{1}{c|}{Carbs$^{(-b)}$} & \multicolumn{1}{c}{Minimum} & \multicolumn{1}{c}{Maximum}
	& \multicolumn{1}{c}{Median} & \multicolumn{1}{c}{Average} & StdDev\\
	\hline
	559 & 215 & 83 & 8.0 & 75.0 & 30.0 & 35.5 & 15.5\\
    563 & 225 & 28 & 5.0 & 84.0 & 31.0 & 33.8 & 18.0\\
    570 & 174 & 39 & 5.0 & 200.0 & 115.0 & 106.1 & 41.5\\
	575 & 297 & 122 & 1.0 & 110.0 & 40.0 & 40.0 & 22.0\\
	588 & 268 & 73 & 2.0 & 60.0 & 20.0 & 22.7 & 14.6\\
	591 & 264 & 60 & 3.0 & 77.0 & 28.0 & 31.5 & 14.1\\
	\hline
	2018 Total & 1443 & 405 & 1.0 & 200.0 & 33.0 & 41.5 & 32.7\\
	\hline
	540 & 234 & 14 & 1.0 & 110.0 & 40.0 & 50.2 & 29.8\\
	544 & 206 & 41 & 1.0 & 175.0 & 60.0 & 68.7 & 36.3\\
	552 & 271 & 25 & 3.0 & 135.0 & 26.0 & 36.7 & 29.3\\
	567 & 207 & 5 & 20.0 & 140.0 & 67.0 & 67.0 & 21.5\\
	584 & 233 & 44 & 15.0 & 78.0 & 60.0 & 54.6 & 11.6\\
	596 & 300 & 277 & 1.0 & 64.0 & 25.0 & 25.1 & 14.0\\
	\hline
	2020 Total & 1451 & 406 & 1.0 & 175.0 & 42.0 & 48.2 & 29.5\\
	\hline
	Combined Total & 2894 & 811 & 1.0 & 200.0 & 39.0 & 44.9 & 31.3\\
	\hline
\end{tabular}
\end{center}
\end{table}

\begin{table}[t]\setlength{\tabcolsep}{4pt}
\begin{center}
\caption{Per subject and total boluses and insulin units statistics: Minimum, Maximum, Median, Average, and Standard Deviation (StdDev). Bolus$^{(\pm c)}$ refers to all bolus events; Bolus$^{(+c)}$ refers to bolus events associated with a meal. Statistics are shown for the 2018 subset, the 2020 subset, and for the entire OhioT1DM dataset.}
\label{tab:boluses}
\begin{tabular}{|crr|rrrrc|}
    \cline{4-8}
    \multicolumn{3}{c|}{} & \multicolumn{5}{c|}{Insulin Per Bolus}\\
	\hline
	Subject & \multicolumn{1}{c}{Bolus$^{(\pm c)}$} & \multicolumn{1}{c|}{Bolus$^{(+c)}$}
	& \multicolumn{1}{c}{Minimum} & \multicolumn{1}{c}{Maximum} & \multicolumn{1}{c}{Median} & \multicolumn{1}{c}{Average} & StdDev\\
	\hline
	559 & 186 & 132 & 0.1 & 9.3 & 3.6 & 3.7 & 1.9\\
    563 & 424 & 197 & 0.1 & 24.7 & 7.8 & 8.0 & 4.2\\
    570 & 1,345 & 132 & 0.2 & 12.1 & 1.3 & 1.8 & 2.1\\
	575 & 271 & 175 & 0.1 & 12.8 & 4.4 & 4.1 & 3.0\\
	588 & 221 & 195 & 0.4 & 10.0 & 3.5 & 4.3 & 2.3\\
	591 & 331 & 204 & 0.1 & 9.4 & 2.9 & 3.1 & 1.8\\
	\hline
	2018 Total & 2,758 & 1,035 & 0.1 & 24.7 & 1.9 & 3.5 & 3.4\\
	\hline
	540 & 521 & 220 & 0.1 & 11.4 & 2.0 & 3.0 & 2.8\\
	544 & 264 & 149 & 0.7 & 22.5 & 5.0 & 6.5 & 4.9\\
	552 & 426 & 246 & 0.1 & 16.0 & 2.8 & 3.9 & 3.3\\
	567 & 366 & 202 & 0.2 & 25.0 & 11.4 & 12.0 & 5.8\\
	584 & 311 & 188 & 0.1 & 16.2 & 9.1 & 7.3 & 3.1\\
	596 & 230 & 0 & 0.2 & 7.6 & 3.3 & 3.0 & 1.5\\
	\hline
	2020 Total & 2,118 & 1,169 & 0.1 & 25.0 & 4.0 & 5.8 & 5.0\\
	\hline
	Combined Total & 4,876 & 2,204 & 0.1 & 25.0 & 2.9 & 4.5 & 4.3\\
	\hline
\end{tabular}
\end{center}
\end{table}

\subsection{From Meals and Bolus Events to Recommendation Examples}
\label{sec:examples}

In all recommendation scenarios, the prediction window ranges between the present time $t$ and the prediction horizon $t + 10 + \tau$. For the carbohydrate or bolus recommendation scenarios, the meal or the bolus is assumed to occur at time $t +10$. For the bolus given carbs scenario, the bolus occurs at time $t+10$ and is followed by a meal at time $t+20$, which matches the pre-processing of the meal data. For evaluation purposes, we set $\tau$ to values between 30 and 90 minutes with a step of 5 minutes, i.e, $\tau \in \{30, 35, 40, ..., 90\}$ for a total of 13 different values. As such, each meal/bolus event in the data results in 13 recommendation examples, one example for each value of $\tau$. While all 13 examples use the same value for the prediction label, e.g., $B_{t + 10}$ for bolus prediction, they will differ in terms of the target BG feature $G_{t + 10 + \tau}$ and the $\tau$ feature, both used directly as input to the FC layers in the architectures shown in Figures~\ref{fig:carbs} and~\ref{fig:bolus}. For the bolus given carbs scenario, the 13 examples are only created when there is a meal that had a bolus delivered 10 minutes prior. Due to the way the data is pre-processed, it is guaranteed that if a meal had a bolus associated with it, the bolus will be exactly 10 minutes before the meal. 

Table \ref{tab:c1_examples} shows the number of {\it inertial} examples for 5 prediction horizons, as well as the total over all 13 possible prediction horizons. Table \ref{tab:c2_examples} shows the number of {\it unrestricted} examples. Since the same number of unrestricted examples are available for every prediction horizon, only the totals are shown. The only exceptions would be if an event was near the end of a subject's data and the prediction horizon $t+10+\tau$ goes past the end of the dataset for some value of $\tau$.

\begin{table}[ht]\setlength{\tabcolsep}{4pt}
\begin{center}
\caption{{\it Inertial} ({\it I}) examples by recommendation scenario and prediction horizon. Carbs$^{(\pm b)}$ refers to all carbohydrate intake events; Carbs$^{(-b)}$ refers to carbohydrate intakes without a bolus.}
\label{tab:c1_examples}
\begin{tabular}{|l|rrrr|rrrr|}
    \cline{2-9}
    \multicolumn{1}{c}{}& \multicolumn{4}{|c|}{Carbs$^{(\pm b)}$ recommendation} & \multicolumn{4}{c|}{Carbs$^{(-b)}$ recommendation}\\
    \hline
    Horizon & Training & Validation & Testing & Total {\it I} & Training & Validation & Testing & Total {\it I}\\
    \hline
    $\tau=30$ & 1,192 & 340 & 331 & 1,863 & 265 & 53 & 40 & 358\\
    $\tau=45$ & 1,156 & 334 & 321 & 1,811 & 255 & 51 & 40 & 346\\ 
    $\tau=60$ & 1,121 & 318 & 315 & 1,754 & 243 & 50 & 40 & 333\\
    $\tau=75$ & 1,057 & 301 & 293 & 1,651 & 226 & 44 & 34 & 304\\ 
    $\tau=90$ & 975 & 279 & 278 & 1,532 & 200 & 40 & 31 & 271\\
    All 13 horizons & 14,343 & 4,103 & 4,007 & 22,453 & 3,100 & 620 & 486 & 4,206\\
    \hline
	\multicolumn{5}{c}{}\\[-1.5ex]
	\cline{2-9}
    \multicolumn{1}{c}{} & \multicolumn{4}{|c|}{Bolus$^{(\pm c)}$ recommendation} & \multicolumn{4}{c|}{Bolus$^{(+ c)}$ recommendation}\\
    \hline
    Horizon & Training & Validation & Testing & Total {\it I} & Training & Validation & Testing & Total {\it I}\\
    \hline
    $\tau=30$ & 461 & 160 & 143 & 764 & 856 & 267 & 271 & 1,394\\
    $\tau=45$ & 416 & 142 & 124 & 682 & 833 & 259 & 258 & 1,350\\
    $\tau=60$ & 368 & 124 & 104 & 596 & 816 & 253 & 249 & 1,318\\
    $\tau=75$ & 303 & 102 & 96 & 501 & 790 & 243 & 243 & 1,276\\
    $\tau=90$ & 271 & 90 & 86 & 447 & 743 & 234 & 229 & 1,206\\
    All 13 horizons & 4,732 & 1,606 & 1,423 & 7,761 & 10,514 & 3,269 & 3,249 & 17,032\\
    \hline
    
\end{tabular}
\end{center}
\end{table}

\begin{table}[ht]\setlength{\tabcolsep}{4pt}
\begin{center}
\caption{{\it Unrestricted} (U) examples by recommendation scenario,  also showing, in the last column, the total number of non-inertial ($U - I$) examples. Carbs$^{(\pm b)}$ refers to all carbohydrate intake events; Carbs$^{(-b)}$ refers to carbohydrate intakes without a bolus.}
\label{tab:c2_examples}
\begin{tabular}{|l|rrrr|r|}
	\hline
	Scenario & Training & Validation & Testing & Total $U$ & Total $U - I$\\
	\hline
	Carbs$^{(\pm b)}$ & 17,937 & 5,106 & 4,943 & 27,986 & 5,533\\
	Carbs$^{(-b)}$  & 4,140 & 853 & 624 & 5,617 & 1,411\\
	Bolus$^{(\pm c)}$ & 19,640 & 6,279 & 6,136 & 32,055 & 24,294\\
	Bolus$^{(+c)}$ & 12,052 & 3,784 & 3,816 & 19,652 & 2,620\\
	\hline
\end{tabular}
\end{center}
\end{table}

For the carbohydrate and bolus given carbs recommendation scenarios, the gap between the number of {\it inertial} and {\it unrestricted} examples is not very large, as most examples qualify as inertial examples. However, in the bolus recommendation scenario, there is a very sizable gap between the number of inertial vs. unrestricted examples. This is because a significant number of boluses are associated with meals, and since these meals are timestamped to be 10 minutes after the bolus, the result is that a bolus at time $t + 10$ will be associated with a meal at time $t + 20$. Therefore, for preprandial boluses at $t + 10$, the meal at time $t + 20$ will prohibit the creation of inertial recommendation examples, because by definition inertial examples do not allow the presence of other events in the prediction window $(t, t + 10 + \tau)$.

\section{Experimental Methodology}
\label{sec:methodology}

For each of the 12 subjects in the dataset, their time series data is split into three sets, as follows:
\begin{itemize}
    \item {\it Testing}: the last 10 days of data.
    \item {\it Validation}: the 10 days of data preceding the testing portion.
    \item {\it Training}: the remainder of the data, around 30 days.
\end{itemize}
The blood glucose, carbs, and insulin values are all scaled to be between $[0, 1]$ by using maximum and minimum values computed over training data. When computing the performance metrics at test time, the predicted values are scaled back to the original range.
The neural architecture is trained to minimize the mean squared error between the actual event (meal or bolus) value recorded in the training data and the estimated value computed by the output node of the fully connected layers in the LSTM models, or by the accumulated forecasts in the N-BEATS architecture. The Adam \cite{kingma:adam} variant of gradient descent is used for training, with the learning rate and mini-batch size being tuned on the validation data. In an effort to avoid overfitting, dropout and early stopping with a patience of 10 epochs are used in all experiments.

Before training a personalized model for a specific subject, a generic model is first pre-trained on the union of all 12 subjects' training data. The generic model is then fine tuned separately for each individual subject, by continuing training on that subject's training data only. The pre-training allows the model parameters to be in a better starting position before fine tuning, allowing faster and better training. The learning rate and batch size are tuned for each subject on their validation data. 
For each subject, the results are aggregated over 10 models that are trained with different seedings of the random number generators.

The metrics used to evaluate the models are the Root Mean Squared Error (RMSE) and the Mean Absolute Error (MAE). Two scores are reported for each of the LSTM-based and N-BEATS-based recommendation models:
\begin{enumerate}
    \item The {\bf $\langle$model$\rangle$.mean} score calculates the average RMSE and MAE on the testing data across the 10 models trained for each subject, and then averages these scores across all subjects.
    \item The {\bf $\langle$model$\rangle$.best} score instead selects for each subject the model that performed best in terms of MAE on the validation data, out of the 10 models trained for that subject. The RMSE and MAE test scores are averaged over all subjects.
\end{enumerate}
Two sets of models were trained for each recommendation scenario: a set of models was trained and evaluated on {\it inertial} examples and a set was trained and evaluated on {\it unrestricted} examples. 


\subsection{Subject Selection for Testing in Each Recommendation Scenario}

While using both the 2018 and 2020 subsets of the OhioT1DM Dataset \cite{ohiot1dm:marling:kdh18, ohiot1dm:marling:kdh20} provides us with data from 12 total subjects, not all 12 can be used in each scenario, due to insufficient examples in their respective development or test subsets. The subjects whose data was used or not at test time are listed below for each scenario, together with a justification:
\begin{itemize}
    \item {\it Carbs$^{(\pm b)}$ Recommendation}: Subjects 567 and 570 were left out at test time. Subject 567 had 0 meal events in the testing portion of their data. Subject 570 frequently used dual boluses; as such, there were very few inertial examples for this subject at all. Of the few inertial examples that were available, 0 were in the testing or validation portions of the data.
    \item {\it Carbs$^{(-b)}$ Recommendation}: Due to the limited number of examples for this scenario, we trained and evaluated models only for the subjects whose data contained at least 50 carb events with no associated bolus. These are subjects 559, 575, 588, and 591. While subject 596 also had a sufficient number of carb events, we discovered that all carbohydrate inputs for their BW were 0. As a consequence of this missing data, it cannot be determined which boluses were used for BGL correction, and which were used to cover meals. Therefore, subject 596 cannot be used in this scenario.
    \item {\it Bolus$^{(\pm c)}$ Recommendation}: Subjects 544 and 567 were left out at test time. Subject 544  had few inertial examples overall, and 0 in the validation portion of the data. This is because the vast majority of bolus events in their data was used in conjunction with a meal. Similar to the carbohydrate recommendation scenario, subject 567 was not used in this scenario because of the lack of meal events in their test data. The missing meal data would make the bolus recommendation results for this subject unrealistic and also indistinguishable between the inertial and unrestricted cases.
    \item {\it Bolus$^{(+c)}$ Recommendation}: Subjects 567, 570, and 596 were left out at test time. As explained for other scenarios above, subject 567 had 0 meals in the test portion of their data. For subject 570, there were 0 inertial examples in the test portion. As explained for the Carbs$^{-b}$ recommendation scenario, due to missing BW data, for subject 596 it cannot be determined which boluses were used for BGL correction, and which were used to cover meals, so their data cannot be used in this scenario, either.
\end{itemize}
Irrespective of which subjects are used at test time, the data from all 12 patients is used for pre-training purposes in each recommendation scenario. Furthermore, the set of subjects stays consistent between the inertial and unrestricted cases for any given recommendation scenario.

\subsection{Evaluating the Impact of Pre-processing of Meals}
\label{sec:pre-evaluation}

To determine the utility of the pre-processing of meals procedure introduced in Section~\ref{sec:pre-processing}, we trained and evaluated N-BEATS-based models for the carbohydrate recommendation scenario Carbs$^{(\pm b)}$ using the original data vs. using the pre-processed data. When training on pre-processed data, we report in Table~\ref{tab:pre-carbs} two development results: when evaluating on all the pre-processed meals in the development data (pre$^+$)  vs. evaluating only on meals that were not added during pre-processing (pre$^-$). The results show that in both cases the pre-processing of meals leads to statistically significant improvements in RMSE and MAE. Pre-processing of meals also benefits the bolus recommendation scenario, as shown in Table~\ref{tab:pre-bolus}. These results can be seen as further evidence of the fact that the meal timestamps recorded in the smartphone interface are unreliable and that meal times should instead be anchored to the bolus timestamps recorded by the BW, as done in the pre-processing procedure.

\begin{table}[ht]
\setlength{\tabcolsep}{4pt}
\caption{Results with pre-processing of meals (pre) vs. original raw data for meal events (raw), for the carbohydrate recommendation scenario Carbs$^{(\pm b)}$ on unrestricted examples. pre$^+$ refers to using all pre-processed meals (shifted original meals and added meals), whereas pre$^-$ does not use meals added by the pre-processing procedure. The symbol $\dagger$ indicates a p-value < 0.03 when using a one-tailed t-test to compare against the results without pre-processing (raw).}
\begin{center}
\label{tab:pre-carbs}
\begin{tabular}{|l|ll|rr|}
    \cline{2-3}
    \multicolumn{1}{c|}{} & \multicolumn{2}{c|}{Pre-processing} &  \multicolumn{2}{c}{} \\
    \cline{2-5}
	\multicolumn{1}{c|}{} & Train & Devel & RMSE & MAE\\
    \hline
    \multirow{2}{*}{N-BEATS.mean} & raw & raw & 13.42 & 10.32\\
	& pre$^+$ & pre$^-$ & $^\dagger$9.38 & $^\dagger$6.59\\
	& pre$^+$ & pre$^+$ & $^\dagger${\bf 8.84} & $^\dagger${\bf 6.16}\\
    \hline
    \multirow{2}{*}{N-BEATS.best} & raw & raw & 12.32 & 9.28\\
    & pre$^+$ & pre$^-$ & $^\dagger$8.48 & $^\dagger$5.90\\
    & pre$^+$ & pre$^+$ & $^\dagger${\bf 8.12} & $^\dagger${\bf 5.53}\\
    \hline
\end{tabular}
\end{center}
\end{table}

\begin{table}[ht]
\setlength{\tabcolsep}{4pt}
\caption{Results with pre-processing of meals (pre) vs. original raw data for meal events (raw), for the Bolus$^{(\pm c)}$ recommendation scenario on unrestricted examples. All meals (shifted or added) are used for the pre-processed data. The symbol $\dagger$ indicates a p-value < 0.01 when using a one-tailed t-test to compare against the results without pre-processing (raw).}
\begin{center}
\label{tab:pre-bolus}
\begin{tabular}{|l|c|rr|}
    \cline{2-4}
	\multicolumn{1}{c|}{} & Pre-processing & RMSE & MAE\\
    \hline
    \multirow{2}{*}{N-BEATS.mean} & raw & 1.85 & 1.41\\
    & pre & $^\dagger${\bf 1.30} & $^\dagger${\bf 0.92}\\
    \hline
    \multirow{2}{*}{N-BEATS.best} & raw & 1.81 & 1.32 \\
    & pre & $^\dagger${\bf 1.22} & $^\dagger${\bf 0.84}\\
    \hline
\end{tabular}
\end{center}
\end{table}

\subsection{Tuning the Architecture and the Hyper-parameters on the Development Data}
\label{sec:development}

Table~\ref{tab:state1} show the results of the LSTM- and N-BEATS-based models, with vs. without using the final state produced by the LSTM$_1$ component as input to the fully connected network. The results show that using the final state from LSTM$_1$ directly as input leads to a substantial improvement for the carbohydrate recommendation scenario Carbs$^{(\pm b)}$, while maintaining a comparable performance for the bolus recommendation scenario. Consequently, in all remaining experiments the architecture is set to use the final state of LSTM$_1$ as input to the FC layers.

\begin{table}[ht]
\caption{Performance of the LSTM- and N-BEATS-based models, with ($+$) and without ($-$) the final state $s_1$ of LSTM$_{1}$ as part of the input to the FC Layers.}
\begin{center}
\label{tab:ablation_lstm1}
\begin{tabular}{|c|c|rr|c|c|c|rr|}
    \cline{3-4} \cline{8-9}
    \multicolumn{2}{c|}{LSTM.mean} & RMSE & MAE & \multicolumn{1}{c}{} & \multicolumn{2}{c|}{N-BEATS.mean} & RMSE & MAE\\
    \cline{1-4} \cline{6-9}
    \multirow{2}{*}{Carbs$^{(\pm b)}$} & $- s_1$ & 10.14 & 7.56 & & \multirow{2}{*}{Carbs$^{(\pm b)}$} & $- s_1$ & 10.27 & 7.58\\
    & $+ s_1$ & {\bf 8.99} & {\bf 6.57} & & & $+ s_1$ & {\bf 8.84} & {\bf 6.16}\\
    \cline{1-4} \cline{6-9}
    \multirow{2}{*}{Bolus$^{(\pm c)}$} & $- s_1$ & {\bf 1.33} & {\bf 0.97} & & \multirow{2}{*}{Bolus$^{(\pm c)}$} & $- s_1$ & 1.33 & {\bf 0.85}\\
    & $+ s_1$ & 1.41 & 1.03 & & & $+ s_1$ & {\bf 1.30} & 0.92\\
    \cline{1-4} \cline{6-9}
\end{tabular}
\end{center}
\label{tab:state1}
\end{table}

In the original N-BEATS model of \citet{oreshkin:nbeats}, the backcast and forecast outputs of each block are produced as the result of two separate fully connected layers. In the block architecture shown in Figures~\ref{fig:carbs},~\ref{fig:bolus}, and~\ref{fig:nbeats} however, the {\it FC Layers} component uses just one final fully connected layer to produce both backcast and forecast values. The results in Table~\ref{tab:splitting} show that, overall, using a joint final layer is competitive or better than using separate layers.

\begin{table}[ht]
\caption{N-BEATS-based model results, with a {\it separate} vs. {\it joint} final fully connected layer for computing backcast and forecast values.}
\begin{center}
\label{tab:ablation_split}
\begin{tabular}{|c|c|c|c|}
	\cline{3-4}
	\multicolumn{2}{c|}{N-BEATS.mean} & RMSE & MAE\\
	\hline
	\multirow{2}{*}{Carbs$^{(\pm b)}$} & {\it separate} & {\bf 8.77} & 6.48\\
	& {\it joint} & 8.84 & {\bf 6.16}\\
	\hline
	\multirow{2}{*}{Bolus$^{(\pm c)}$} & {\it separate} & 1.32 & 0.94\\
	& {\it joint} & {\bf 1.30} & {\bf 0.92}\\
	\hline
\end{tabular}
\end{center}
\label{tab:splitting}
\end{table}

For each prediction scenario, the hyper-parameters for both the LSTM-based and N-BEATS-based models were tuned on development data. The inertial and unrestricted models are tuned independent of each other. The learning rate was tuned by monitoring the learning curves, using values between 0.0002 \cite{rubin_falcone:nbeats_bgl} and 0.1. After multiple experiments, a fixed learning rate of 0.001 was observed to give the best results on development data in all scenarios. The number of blocks in N-BEATS, the number of FC layers in the LSTM, and the dropout rate were then tuned in that order. The number of N-BEATS blocks was selected from \{1, ..., 10\}, the number of layers was selected from \{1, 2, 3, 4, 5\}, whereas the dropout rate was tuned with values from \{0, 0.1, 0.2, 0.3, 0.4 0.5\}. The tuned values are shown in Table~\ref{tab:hyper-lstm} for the LSTM models and Table~\ref{tab:hyper-nbeats} for the N-BEATS models. Overall, the LSTM-based models worked best with only 2 or 3 fully connected layers in all scenarios, whereas the N-BEATS-based models worked best with 4 or 5 fully connected layers. The tuned number of blocks in the N-BEATS-based models varied between 3 and 5, depending on the scenario and the unrestricted vs. inertial case. The tuned dropout rates varied a lot between scenarios for the LSTM-based models, with rates ranging from 0 to 0.5, whereas the tuned rates for N-BEATS-based models varied between 0.2 and 0.5.

\begin{table}[ht]
\caption{Tuned hyper-parameters for the LSTM-based models.}
\label{tab:hyper-lstm}
\begin{center}
\begin{tabular}{|c|c|c|c|}
\cline{3-4}
\multicolumn{2}{l}{} & \multicolumn{2}{|c|}{Hyper-Parameters}\\
\hline
Scenario & Examples & FC Layers & Dropout\\
\hline
\multirow{2}{*}{Carbs$^{(\pm b)}$} & Inertial & 3 & 0.1\\
& Unrestricted & 3 & 0.1\\
\hline
\multirow{2}{*}{Bolus$^{(\pm c)}$} & Inertial & 3 & 0.0\\
& Unrestricted & 2 & 0.3\\
\hline
\multirow{2}{*}{Bolus$^{(+c)}$} & Inertial & 2 & 0.2\\
& Unrestricted & 2 & 0.5\\
\hline
\end{tabular}
\end{center}
\end{table}

\begin{table}[ht]
\caption{Tuned hyper-parameters for the N-BEATS-based models.}
\label{tab:hyper-nbeats}
\begin{center}
\begin{tabular}{|c|c|c|c|c|}
\cline{3-5}
\multicolumn{2}{l}{} & \multicolumn{3}{|c|}{Hyper-Parameters}\\
\hline
Scenario & Examples & Blocks & FC Layers & Dropout\\
\hline
\multirow{2}{*}{Carbs$^{(\pm b)}$} & Inertial & 5 & 2 & 0.3\\
& Unrestricted & 3 & 3 & 0.3\\
\hline
\multirow{2}{*}{Bolus$^{(\pm c)}$} & Inertial & 5 & 4 & 0.2\\
& Unrestricted & 4 & 4 & 0.2\\
\hline
\multirow{2}{*}{Bolus$^{(+c)}$} & Inertial & 5 & 4 & 0.5\\
& Unrestricted & 3 & 5 & 0.2\\
\hline
\end{tabular}
\end{center}
\end{table}

The size of the LSTM state was tuned to 32, whereas the size of each fully connected layer was tuned to 64, which is substantially smaller than the hidden size of 512 used in the original N-BEATS model \cite{oreshkin:nbeats}. For the carbohydrates without bolus scenario Carbs$^{(-b)}$, due to the much smaller number of examples, we reduced the number of units in the LSTM networks and fully connected layers by a factor of 2. The same hyper-parameters that were tuned on the general carbohydrate recommendation scenario Carbs$^{(\pm b)}$ were used for Carbs$^{(-b)}$.

\section{Experimental Results}
\label{sec:results}

\begin{table*}[t]\setlength{\tabcolsep}{4pt}
\caption{Results for each recommendation scenario, for both classes of examples. The simple $\dagger$ indicates a p-value < 0.05 when using a one-tailed t-test to compare against the baseline results; the double $\ddagger$ indicates statistical significance for comparison against the baselines as well as against the competing neural method; the $\uparrow$ indicates significant with respect to the Global Average baseline only.}
\begin{center}
\label{tab:results}
\begin{tabular}{|l|rr|rr|}

   	\cline{2-5}
	\multicolumn{1}{c}{} & \multicolumn{2}{|c|}{Inertial} & \multicolumn{2}{c|}{Unrestricted}\\
	\hline
	Carbs$^{(\pm b)}$ recommendation & RMSE & MAE & RMSE & MAE\\
	\hline
	Global Average & 20.90 & 17.30 & 20.68 & 17.10\\
	ToD Average & 20.01 & 15.78 & 19.82 & 15.68\\
	\hline
	LSTM.mean & 11.55 & 7.81 & 10.99 & 7.40\\
	LSTM.best & 10.95 & 7.50 & 10.50 & 7.31\\
	\hline
	N-BEATS.mean & $^\ddagger${\bf 9.79} & $^\ddagger${\bf 6.45} & 10.34 & 7.04\\
	N-BEATS.best & 9.92 & 6.56 & $^\dagger${\bf 10.07} & $^\dagger${\bf 6.75}\\
	\hline
	\multicolumn{5}{c}{}\\[-1.5ex]
	\cline{2-5}
	\multicolumn{1}{c}{} & \multicolumn{2}{|c|}{Inertial} & \multicolumn{2}{c|}{Unrestricted}\\
	\hline
	Carbs$^{(-b)}$ recommendation & RMSE & MAE & RMSE & MAE\\
	\hline
	Global Average & 15.92 & 13.71 & 14.66 & 12.19\\
	ToD Average & 15.55 & 13.45 & 14.27 & 11.93\\
	\hline
	LSTM.mean & 14.02 & 11.47 & 14.70 & 12.27\\
	LSTM.best & 13.75 & 10.92 & 14.94 & 12.57\\
	\hline
	N-BEATS.mean & {\bf 13.76} & {\bf 11.42} & $^\uparrow${\bf 13.69} & $^\uparrow${\bf 11.09}\\
	N-BEATS.best & 14.52 & 11.78 & 14.17 & 11.47\\
	\hline
	\multicolumn{5}{c}{}\\[-1.5ex]
   	\cline{2-5}
	\multicolumn{1}{c}{} & \multicolumn{2}{|c|}{Inertial} & \multicolumn{2}{c|}{Unrestricted}\\
	\hline
	Bolus$^{(\pm c)}$ recommendation & RMSE & MAE & RMSE & MAE\\
	\hline
	Global Average & 2.40 & 2.13 & 2.84 & 2.30\\
	ToD Average & 2.21 & 1.86 & 2.71 & 2.17\\
	\hline
	LSTM.mean & 1.75 & 1.35 & 1.53 & 1.10\\
	LSTM.best & 1.70 & 1.30 & 1.50 & 1.05\\
	\hline
	N-BEATS.mean & $^\dagger${\bf 1.56} & $^\ddagger${\bf 1.20} & $^\dagger${\bf 1.49} & 1.04\\
	N-BEATS.best & 1.65 & 1.26 & 1.51 & $^\dagger${\bf 1.03}\\
	\hline
	\multicolumn{5}{c}{}\\[-1.5ex]
	\cline{2-5}
	\multicolumn{1}{c}{} & \multicolumn{2}{|c|}{Inertial} & \multicolumn{2}{c|}{Unrestricted}\\
	\hline
	Bolus$^{(+c)}$ recommendation & RMSE & MAE & RMSE & MAE\\
	\hline
	Global Average & 3.00 & 2.35 & 3.04 & 2.39\\
	ToD Average & 2.87 & 2.21 & 2.90 & 2.25\\
	\hline
	LSTM.mean & 1.02 & 0.73 & {\bf 1.00} & 0.73\\
	LSTM.best & 0.94 & 0.67 & $^\dagger${\bf 1.00} & $^\dagger${\bf 0.72}\\
	\hline
	N-BEATS.mean & 0.89 & 0.65 & 1.11 & 0.82\\
	N-BEATS.best & $^\dagger${\bf 0.85} & $^\dagger${\bf 0.61} & 1.06 & 0.78\\
	\hline
\end{tabular}
\end{center}
\end{table*}

Table~\ref{tab:results} shows the results for the two baselines and the two neural architectures: the LSTM-based (Figures~\ref{fig:carbs} and ~\ref{fig:bolus}) and the N-BEATS-based (Figure~\ref{fig:nbeats}). Across all scenarios and for both example classes, the neural models outperform both baselines, often by a wide margin. Furthermore, the N-BEATS-based models outperform their LSTM-based counterparts across all evaluations with inertial examples, which are the ones with the most practical utility. In general, there is little difference between the best model scores and the average model scores, which means that the model performance is relatively stable with respect to the random initialization of the network parameters.

For the prediction of carbohydrates without an associated bolus scenario Carbs$^{(-b)}$, the improvement brought by the two neural models over the two baselines was less substantial, which is to be expected for two reasons. First, the baselines do much better in this scenario than in the more general carbohydrate recommendation scenario Carbs$^{(\pm b)}$ because most of the carb intakes are relatively small, e.g. hypo correction events where subjects are advised to eat a fixed amount of carbohydrate. Second, and most importantly, the number of training carbohydrate events and their associated examples in the Carbs$^{(-b)}$ scenario is much smaller than in the Carbs$^{(\pm b)}$ scenario (Table~\ref{tab:meals}), which makes ML models much less effective.

In all experiments reported so far, one model was trained for all prediction horizons, using the value of $\tau \in \{30, 35, ..., 90\}$ as an additional input feature. This global model was then tested on examples from all prediction horizons. To determine if transfer learning happens among different prediction horizons, for each value of $\tau \in \{30, 45, 60, 75, 90\}$ at test time, we compare the performance of the globally trained model vs. the performance of a model trained only on examples for that particular prediction horizon, using inertial examples for both. We chose the inertial case for this experiment because it corresponds better to the intended use of a carbohydrate or bolus recommendation system. Furthermore, we experiment only with the N-BEATS-based model because of its better performance in the inertial case. The results in Table \ref{tab:transfer_time} show transfer learning clearly happening for the carbohydrate recommendation Carbs$^{(\pm b)}$ and bolus given carbs recommendation Bolus$^{(+c)}$ scenarios, where the models trained on all prediction horizons outperform those trained only on a specific prediction horizon when evaluated on that prediction horizon. For the bolus recommendation scenario Bolus$^{(-c)}$ (i.e. Bolus$^{(\pm c)}$ inertial) the results were mixed, with transfer learning being clear only for the short $\tau = 30$ time horizon. Transfer learning results for the Carbs$^{(-b)}$ scenario are not calculated due to the lack of a sufficient number of training examples for each prediction horizon.

\begin{table}[ht]
\setlength{\tabcolsep}{1.75pt}
\caption{Comparison between models trained on all prediction horizons vs. one prediction horizon $\tau$, when evaluated on the prediction horizon $\tau$. The symbol $\dagger$ indicates a p-value < 0.05 when using a one-tailed t-test to compare against the one prediction horizon results.}
\begin{center}
\label{tab:transfer_time}
\begin{tabular}{|c|c|rr|rr|rr|rr|rr|rr|rr}
    \cline{3-14}
    \multicolumn{2}{c|}{} & \multicolumn{12}{c|}{Carbs$^{(\pm b)}$ recommendation}\\
    \cline{3-14}
    \multicolumn{2}{c|}{} & \multicolumn{2}{c|}{$\tau=30$} & \multicolumn{2}{c|}{$\tau=45$} & \multicolumn{2}{c|}{$\tau=60$} & \multicolumn{2}{c|}{$\tau=75$} & \multicolumn{2}{c|}{$\tau=90$} & \multicolumn{2}{c|}{Average}\\
    \cline{2-14}
     \multicolumn{1}{c|}{}& Trained on & \multicolumn{2}{c|}{\scriptsize RMSE MAE} & \multicolumn{2}{c|}{\scriptsize RMSE MAE} & \multicolumn{2}{c|}{\scriptsize RMSE MAE} & \multicolumn{2}{c|}{\scriptsize RMSE MAE} & \multicolumn{2}{c|}{\scriptsize RMSE MAE} & \multicolumn{2}{c|}{\scriptsize RMSE MAE} \\
    \hline
    \multirow{2}{*}{N-BEATS.mean} & One $\tau$ & {\bf 9.74} & 6.72 & 10.24 & 6.89 & 10.06 & 6.85 & 10.52 & 7.19 & 9.82 & 6.73 & 10.08 & 6.88\\
    & All $\tau$ & 9.96 & {\bf 6.57} & {\bf 9.98} & {\bf 6.56} & {\bf 9.84} & {\bf 6.50} & $^\dagger${\bf 9.55} & $^\dagger${\bf 6.30}& {\bf 9.37} & {\bf 6.22} & {\bf 9.74} & {\bf 6.43}\\
    \hline
    \multirow{2}{*}{N-BEATS.best} & One $\tau$ & 9.92 & 6.70 & 10.39 & 6.90 & 10.21 & 6.88 & 10.62 & 7.18 & 9.92 & 6.66 & 10.21 & 6.86\\
    & All $\tau$ & {\bf 9.84} & {\bf 6.50} & {\bf 9.94} & {\bf 6.56} & {\bf 10.02} & {\bf 6.57} & {\bf 9.76} & $^\dagger${\bf 6.34} & {\bf 9.43} & {\bf 6.08} & {\bf 9.80} & {\bf 6.41}\\
    \hline
    
    \multicolumn{14}{c}{}\\[-1.5ex]

    \cline{3-14}
    \multicolumn{2}{c|}{} & \multicolumn{12}{c|}{Bolus$^{(-c)}$ recommendation}\\
    \cline{3-14}
    \multicolumn{2}{c|}{} & \multicolumn{2}{c|}{$\tau=30$} & \multicolumn{2}{c|}{$\tau=45$} & \multicolumn{2}{c|}{$\tau=60$} & \multicolumn{2}{c|}{$\tau=75$} & \multicolumn{2}{c|}{$\tau=90$} & \multicolumn{2}{c|}{Average}\\
    \cline{2-14}
     \multicolumn{1}{c|}{}& Trained on & \multicolumn{2}{c|}{\scriptsize RMSE MAE} & \multicolumn{2}{c|}{\scriptsize RMSE MAE} & \multicolumn{2}{c|}{\scriptsize RMSE MAE} & \multicolumn{2}{c|}{\scriptsize RMSE MAE} & \multicolumn{2}{c|}{\scriptsize RMSE MAE} & \multicolumn{2}{c|}{\scriptsize RMSE MAE} \\
    \hline
    \multirow{2}{*}{N-BEATS.mean} & One $\tau$ & 1.82 & 1.42 & {\bf 1.57} & {\bf 1.24} & 1.51 & 1.24 & {\bf 1.37} & {\bf 1.10} & 1.40 & 1.17 & 1.53 & 1.23\\
    & All $\tau$ & {\bf 1.75} & {\bf 1.33} & 1.61 & {\bf 1.24} & {\bf 1.47} & $^\dagger${\bf 1.17} & 1.38 & {\bf 1.10} & {\bf 1.28} & $^\dagger${\bf 1.03} & {\bf 1.50} & $^\dagger${\bf 1.17}\\
    \hline
    \multirow{2}{*}{N-BEATS.best} & One $\tau$ & 1.77 & 1.37 & {\bf 1.54} & {\bf 1.21} & {\bf 1.51} & {\bf 1.23} & {\bf 1.38} & {\bf 1.10} & {\bf 1.34} & {\bf 1.11} & {\bf 1.51} & {\bf 1.20}\\
    & All $\tau$ & {\bf 1.72} & {\bf 1.28} & 1.75 & 1.33 & 1.58 & {\bf 1.23} & 1.45 & 1.12 & 1.44 & 1.13 & 1.59 & 1.22\\
    \hline
    
    \multicolumn{14}{c}{}\\[-1.5ex]

    \cline{3-14}
    \multicolumn{2}{c|}{} & \multicolumn{12}{c|}{Bolus$^{(+c)}$ recommendation}\\
    \cline{3-14}
    \multicolumn{2}{c|}{} & \multicolumn{2}{c|}{$\tau=30$} & \multicolumn{2}{c|}{$\tau=45$} & \multicolumn{2}{c|}{$\tau=60$} & \multicolumn{2}{c|}{$\tau=75$} & \multicolumn{2}{c|}{$\tau=90$} & \multicolumn{2}{c|}{Average}\\
    \cline{2-14}
     \multicolumn{1}{c|}{}& Trained on & \multicolumn{2}{c|}{\scriptsize RMSE MAE} & \multicolumn{2}{c|}{\scriptsize RMSE MAE} & \multicolumn{2}{c|}{\scriptsize RMSE MAE} & \multicolumn{2}{c|}{\scriptsize RMSE MAE} & \multicolumn{2}{c|}{\scriptsize RMSE MAE} & \multicolumn{2}{c|}{\scriptsize RMSE MAE} \\
    \hline
    \multirow{2}{*}{N-BEATS.mean} & One $\tau$ & 0.98 & 0.73 & 0.91 & 0.69 & 0.91 & 0.69 & 0.95 & 0.74 & 0.93 & 0.72 & 0.94 & 0.71\\
    & All $\tau$ & {\bf 0.95} & {\bf 0.68} & {\bf 0.87} & {\bf 0.65} & {\bf 0.86} & {\bf 0.65} & $^\dagger${\bf 0.87} & $^\dagger${\bf 0.65} & $^\dagger${\bf 0.86} & $^\dagger${\bf 0.64} & $^\dagger${\bf 0.88} & $^\dagger${\bf 0.65}\\
    \hline
    \multirow{2}{*}{N-BEATS.best} & One $\tau$ & {\bf 0.94} & 0.69 & 0.91 & 0.69 & 0.92 & 0.68 & 0.93 & 0.71 & 0.91 & 0.70 & 0.92 & 0.69\\
    & All $\tau$ & {\bf 0.94} & {\bf 0.66} & {\bf 0.84} & $^\dagger${\bf 0.62} & $^\dagger${\bf 0.82} & $^\dagger${\bf 0.59} & $^\dagger${\bf 0.82} & $^\dagger${\bf 0.61} & $^\dagger${\bf 0.83} & $^\dagger${\bf 0.61} & $^\dagger${\bf 0.85} & $^\dagger${\bf 0.62}\\
    \hline
\end{tabular}
\end{center}
\end{table}

\section{Conclusion and Future Directions}
\label{sec:conclusion}

We have introduced a general LSTM-based neural architecture, composed of two chained LSTMs and a fully connected network, with the purpose of training  models for making recommendations with respect to any type of quantitative events that may impact blood glucose levels, in particular, carbohydrate amounts and bolus insulin dosages. A deep residual N-BEATS-based architecture was also developed, using the chained LSTMs as a component in its block structure. Experimental evaluations show that the proposed neural architectures substantially outperform a global average baseline as well as a time of day dependent baseline, with the N-BEATS-based models outperforming the LSTM-based counterparts in all evaluations with inertial examples. The trained models are shown to benefit from transfer learning and from a pre-processing of meal events that anchors their timestamps shortly after their corresponding boluses. Overall, these results suggest that the proposed recommendation approaches hold significant promise for easing the complexity of self-managing blood glucose levels in type 1 diabetes. Potential future research directions include investigating the proposed pre-processing of carbohydrate events for blood glucose level prediction and exploring the utility of the two neural architectures for recommending exercise.

\authorcontributions{Jeremy Beauchamp contributed with respect to conceptualization, investigation, methodology, software, validation, writing - original draft, and writing - review \& editing. Razvan Bunescu contributed with respect to conceptualization, formal analysis, funding acquisition, methodology, project administration, resources, software, supervision, validation, visualization, writing - original draft, and writing - review \& editing.
Cindy Marling contributed with respect to conceptualization, data curation, investigation, funding acquisition, and writing - review \& editing.
Zhongen Li contributed with respect to methodology, software, and validation.
Chang Liu contributed with respect to resources and supervision.}


\funding{This research was funded by  grant 1R21EB022356 from the National Institutes of Health (NIH).}

\acknowledgments{Conversations with Josep Vehi helped shape the research directions presented herein.  The contributions of physician collaborators Frank Schwartz, MD, and Amber Healy, DO, are gratefully acknowledged.  We would also like to thank the anonymous people with type 1 diabetes who provided their blood glucose, insulin, and meal data.}

\conflictsofinterest{The authors declare no conflict of interest. The funders had no role in the design of the study, in the collection, analyses, or interpretation of data, in the writing of the manuscript, or in the decision to publish the results.} 






\externalbibliography{yes}
\bibliography{sensor21}



\end{document}